\documentclass[letterpaper, 10 pt, conference]{ieeeconf}  % Comment this line out if you need a4paper

\IEEEoverridecommandlockouts                              % This command is only needed if 
\usepackage{graphics} % for pdf, bitmapped graphics files
\usepackage{graphicx} % for pdf, bitmapped graphics files
\usepackage{epsfig} % for postscript graphics files
\usepackage{times} % assumes new font selection scheme installed
\usepackage{amsmath} % assumes amsmath package installed
\usepackage{amssymb}  % assumes amsmath package installed
\usepackage{lipsum}
\usepackage{float}
\usepackage{booktabs}
\usepackage{cuted}

\usepackage{stfloats}
\usepackage{afterpage}
\usepackage{caption}

\usepackage[font=normalsize]{subfig}
\title{\LARGE \bf Vision-guided Autonomous Dual-arm Extraction Robot for Bell Pepper Harvesting}

\author{Kshitij Madhav Bhat$^{\dagger 1}$, Tom Gao$^{\dagger 1}$, Abhishek Mathur$^{\dagger 1}$, Rohit Satishkumar$^{\dagger 1}$, \\ Francisco Yandun$^{1}$, Dominik Bauer$^{1}$, Nancy Pollard$^{1}$%
\thanks{$^{\dagger}$Equal contribution.}% 
\thanks{$^{1}$All authors are with the Robotics Institute, Carnegie Mellon University, Pittsburgh, PA. 
{\tt\small \{kmadhavb, zimingg, armathur, rsatishk, fyandun, dominikb, nsp\}@cs.cmu.edu}}%
}

\begin{document}

\maketitle
\begin{strip}
\vspace{-2cm}
    \centering
    \includegraphics[width=\textwidth]{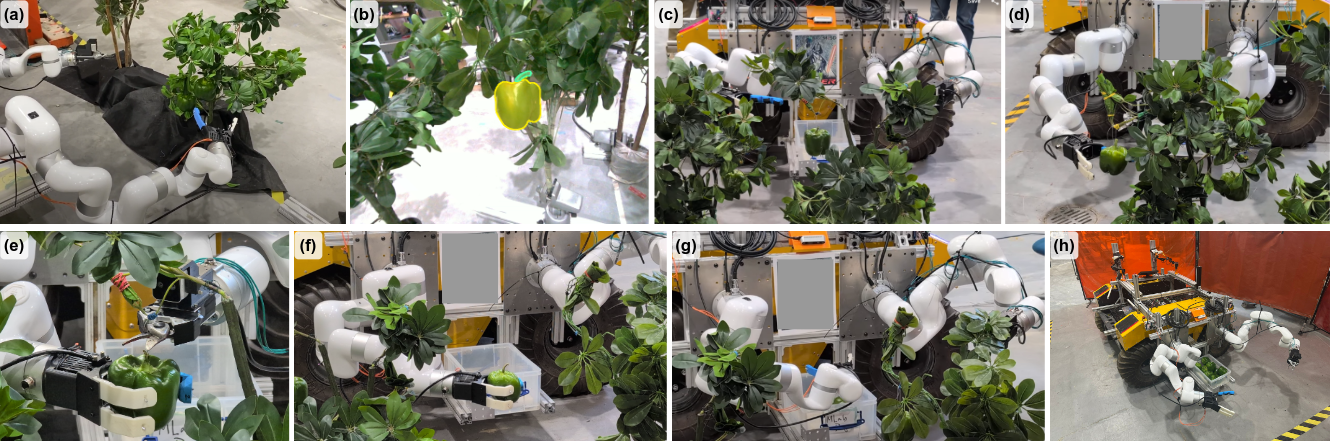}
    \captionof{figure}{VADER dual-arm harvesting approach. 
    \textbf{(a)} Two RGBD cameras scan the scene for peppers. 
    \textbf{(b)} Peppers are detected and their poses are estimated. 
    \textbf{(c)} The gripper and cutter manipulators plan coordinated grasping and cutting motions respectively. 
    \textbf{(d)} The gripper approaches the target pepper. 
    \textbf{(e)} The cutter aligns with the peduncle. 
    \textbf{(f)} The pepper is separated from the plant. 
    \textbf{(g)} The gripper transfers the harvested pepper to the onboard storage bin. 
    \textbf{(h)} Complete system overview of VADER mounted on a mobile platform.}
        \label{fig:vader_system}
\end{strip}

%%%%%%%%%%%%%%%%%%%%%%%%%%%%%%%%%%%%%%%%%%%%%%%%%%%%%%%%%%%%%%%%%%%%%%%%%%%%%%%%
% This is the "Magic" part that fills both columns under the image
\begin{abstract}
Agricultural robotics has emerged as a critical solution to the labor shortages and rising costs associated with manual crop harvesting. Bell pepper harvesting, in particular, is a labor-intensive task, accounting for up to 50\% of total production costs. While automated solutions have shown promise in controlled greenhouse environments, harvesting in unstructured outdoor farms remains an open challenge due to environmental variability and occlusion. This paper presents VADER (Vision-guided Autonomous Dual-arm Extraction Robot), a dual-arm mobile manipulation system designed specifically for the autonomous harvesting of bell peppers in outdoor environments. The system integrates a robust perception pipeline coupled with a dual-arm planning framework that coordinates a gripping arm and a cutting arm for extraction. We validate the system through trials in various realistic conditions, demonstrating a harvest success rate exceeding 60\% with a cycle time of under 100 seconds per fruit, while also featuring a teleoperation fail-safe based on the GELLO teleoperation framework to ensure robustness. To support robust perception, we contribute a hierarchically structured dataset of over 3,200 images spanning indoor and outdoor domains, pairing wide-field scene images with close-up pepper images to enable a coarse-to-fine training strategy from fruit detection to high-precision pose estimation. The code and dataset will be made publicly available upon acceptance.
\end{abstract}

% Force the Introduction to start immediately
\section{INTRODUCTION}
The agricultural industry is currently facing a growing crisis driven by a global shortage of skilled labor and increasing operational costs. Harvesting, traditionally a manual and labor-intensive process, represents a substantial portion of agricultural expenses. In the context of bell pepper production, harvesting alone can account for up to 50\% of the total production cost \cite{panq2024}. Consequently, the development of autonomous robotic systems capable of selective harvesting has become a focal point for both academic research and industrial application.

Existing research in automated harvesting has primarily focused on greenhouse environments, where conditions such as lighting, plant spacing, and terrain are structured and predictable. However, a significant proportion of small to mid-sized farms cultivate peppers in outdoor fields, where unstructured environments pose severe challenges to robotic perception and manipulation. Most current solutions \cite{panq2024, bac2017performance, lehnert2017autonomous, arad2020development, kim2024} often rely on single-manipulator configurations that struggle with the dexterity required to detach fruits without damaging the plant or the produce. Furthermore, the high capital investment required to transition from open-field farming to greenhouse infrastructure renders many existing automated solutions inaccessible to a large segment of growers. Compounding these challenges, existing perception datasets \cite{smitt2021pathobot, smitt2023pag} for bell pepper harvesting focus primarily on the pepper fruit annotations without giving much attention to the peduncle annotations. This gap limits the 
development of accurate pose estimation and cutting-point localization 
methods, especially for multi-arm systems where precise peduncle 
localization directly determines cut success. 

To address these limitations, we propose VADER, a dual-arm autonomous harvesting system mounted on a rugged mobile base, designed to bridge the gap between greenhouse automation and outdoor field requirements. VADER mimics human bimanual dexterity by utilizing separate end-effectors for gripping and cutting, thereby decoupling the grasp stability problem from the detachment task. We also contribute a dataset of images with both coarse (pepper fruit only) and fine (fruit + peduncle) annotations of peppers in order to enable more accurate harvesting of peppers.

The primary contributions of this paper are as follows:

\begin{itemize}
    \item \textbf{An autonomous dual-arm mobile pepper harvesting framework with failsafe teleoperation}: We present a unified framework integrating two 7-DOF manipulators on a Clearpath Warthog mobile base, designed for autonomous coordinated grasp-and-cut operations.
    % : We present a unified framework integrating two 7-DOF manipulators on a Clearpath Warthog mobile base, specifically designed for coordinated grasp-and-cut operations. This architecture combines autonomous dual-arm coordination for improved harvesting reliability in unstructured settings with a seamless, low-latency teleoperation fallback. By utilizing kinematic mimicry \cite{wu2024GELLO}, the system ensures a failsafe mechanism for handling complex edge cases that exceed the current capabilities of autonomous planning.

    \item \textbf{A comprehensive dataset for perception}: To address the lack of high-fidelity data for automated harvesting, we introduce a robust perception dataset over 3,200 annotated images with coarse-grained (pepper-only) and fine-grained (pepper and peduncle) segments.
    
    % : To address the lack of high-fidelity data for automated harvesting, we introduce a robust perception dataset over 3,250 annotated images. This collection is uniquely categorized into coarse-grained (pepper-only) and fine-grained (pepper and peduncle) segments. Our data spans both controlled indoor environments (2,118 images) and challenging outdoor scenarios (1,140 images), covering a spectrum of lighting conditions from direct midday sun to evening sunset.

    \item \textbf{Experimental benchmarking of the dual-arm approach}:  We provide an empirical evaluation of our end to end harvesting framework through extensive indoor trials.
    % : We provide an empirical evaluation of our end to end harvesting framework through extensive indoor trials. Our results demonstrate an autonomous harvest success rate of over 80\% in best-case scenarios and show robust operation under sensor occlusion. Furthermore, we demonstrate that the GELLO-based teleoperation provides a recovery mechanism for complex failure cases.
\end{itemize}

\section{RELATED WORK}
\subsection{Autonomous Bell Pepper Harvesting}
Agricultural robotics has seen significant advancements in recent years in common horticultural tasks including pruning \cite{silwal2022, pruning}, and selective harvesting \cite{strawberry, aubregine, tomato, quadruped}. Among these, non-destructive selective harvesting remains particularly challenging, requiring robust perception, environment-aware motion planning, precise peduncle cutting, and gentle fruit handling. Sweet peppers are among the most difficult crops to harvest autonomously due to severe leaf occlusions, variation in fruit shape and pose, and the need to cut the peduncle without damaging the fruit or plant. Several single-arm systems have been developed for this task. The CROPS project \cite{bac2017performance} used a 9-DOF manipulator with two end-effector designs, achieving only 6\% harvest success in unmodified crop and 33\% in simplified conditions, with a cycle time of 94 s. Harvey \cite{lehnert2017autonomous, lehnert2018sweet} improved upon this with a suction cup and oscillating blade connected by a magnetic decoupling mechanism, reaching 76.5\% success in modified crop. SWEEPER \cite{arad2020development} demonstrated harvesting in a commercial greenhouse with a vibrating knife end-effector. More recently, Pan et al. \cite{panq2024} developed a two-finger parallel gripper with tactile sensing and a swing-cutting module in a plant factory. These single-arm approaches require the same arm to both hold and cut the fruit, necessitating complex end-effector mechanisms. Pan et al. relies on a passive fruit recovery device, leading to harvest failures from plant shaking or the fruit missing the device. Single-arm methods also lack the ability to actively manage occlusions during manipulation. Kim et al. \cite{kim2024} recently proposed an imitation learning approach for outdoor pepper harvesting using a single-arm shear-gripper. While this work is notable as the first to target unprotected outdoor fields, its single-arm shear-gripper design couples cutting and grasping into a single mechanism, limiting the ability to independently optimize each operation, and the system does not explicitly detect or localize the peduncle, contributing to frequent missed cuts.

Multi-arm setups address some of these limitations. Lenz et al. \cite{lenz2024hortibot} proposed HortiBot, a three-arm system with a dedicated perception arm carrying stereo cameras, a pneumatic soft gripper for grasping, and a scissor-based cutter. While this approach enables active perception and dual-arm manipulation, it relies on a pneumatic actuation system that adds mechanical complexity, and its evaluation was limited to a static lab setup rather than a mobile field platform. In contrast, our work presents a dual-arm mobile manipulation system with a soft, tendon-actuated 3D-printed gripper that gently conforms to the fruit without requiring force sensing or pneumatic systems. Dual RGB-D cameras mounted on each end-effector provide redundant perception that mitigates single-viewpoint occlusion failures. The gripper securely holds the pepper throughout cutting and transport, eliminating harvest failures due to plant disturbances common in outdoor settings.
% Across the literature, existing systems predominantly target red or yellow peppers in greenhouse or indoor environments, leaving a significant gap for green pepper harvesting in outdoor settings where uncontrolled lighting and color similarity with foliage make perception far more challenging.

\subsection{Agricultural Perception Datasets}
The efficiency of robotic harvesting is deeply influenced by the quality of underlying training data. Within the domain of bell pepper harvesting, existing datasets fail to balance scale with the detailed, paired fruit-peduncle annotations required for accurate pose estimation and autonomous harvesting. Early benchmarks like \textit{BUP19} \cite{halstead2020fruit} introduced detection datasets captured in natural conditions, while the subsequent \textit{BUP20} collection from the \textit{PATHoBot} platform \cite{smitt2021pathobot} scaled this effort considerably. Recent extensions of \textit{BUP20} \cite{smitt2023pag} provide panoptic segmentation predictions for nearly 18,000 images. While these large-scale spatial-temporal datasets are valuable for crop monitoring and yield estimation, they are restricted to glasshouse environments and rely on automated model-based predictions rather than human-verified annotations. Specialized datasets such as \cite{montoya2021sweet} include peduncle labels but exhibit minimal variation in fruit scale and contain fewer than 800 human-verified images. More broadly, across the literature, existing systems and datasets predominantly target red or yellow peppers in greenhouse or indoor environments, leaving a significant gap for green pepper harvesting in outdoor settings where uncontrolled lighting and color similarity with foliage make both perception and segmentation considerably more challenging. Our dataset addresses these limitations by providing over 3,200 curated images balanced across indoor and outdoor domains, featuring human-verified coarse and fine masks for peppers along with their peduncles. This paired annotation provides the high-fidelity fruit and peduncle localization required for the coordinated grasp-and-cut operations that distinguish our dual-arm harvesting system.
\begin{figure*}[ht]
    \centering
    \includegraphics[trim={0, 0.3cm, 0, 0.2cm}, clip, width=\linewidth]{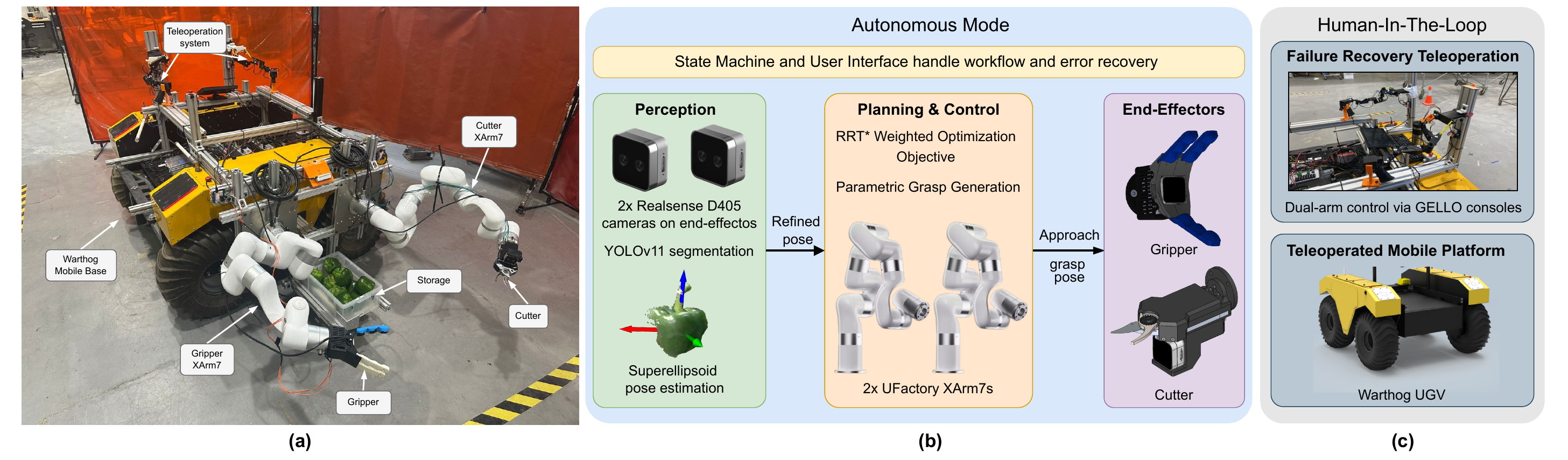}
    \caption{{Overview of VADER. \textbf{(a)} A Warthog UGV base carrying two UFactory XArm7 manipulators with a tendon-actuated 3-fingered soft gripper and peduncle cutter, which are both composed of 3D printed parts. \textbf{(b)} Autonomous pipeline: palm-mounted RealSense D405 cameras feed YOLOv11 segmentation and superellipsoid pose estimation into a collision-aware motion planner, coordinating gripper and cutter for reliable harvest and transfer to storage. \textbf{(c)} GELLO-based dual-arm teleoperation enables failure recovery.}}
    \label{fig:full_overview}
\end{figure*}

\section{METHODS}

\subsection{System Overview}

VADER is comprised of four main subsystems: 1) the Warthog mobile base, 2) two UFactory XArm7 (7 DoF) manipulators with attached end effectors, 3) GELLO teleoperation consoles, and 4) supporting electrical, computing, and user interface components. The Warthog mobile base provides mounting points and onboard regulated power for the system. Since pepper farms arrange plants in rows and the Warthog is differential drive, the manipulators are mounted on the side of the platform to maximize reachability, spaced 50 cm apart to maximize their shared workspace and provide clearance for depositing harvested peppers into the storage bin between the arms. Each XArm7 is equipped with a palm-mounted RealSense D405 RGB-D camera and one of two custom end effectors: a tendon-actuated 3D-printed TPU soft gripper \cite{bauer2023soft} for adaptive grasping of irregular geometries, and a single-motor cutter designed to exert shear force for peduncle severance. This dual-camera configuration provides complementary views that mitigate single-viewpoint occlusion from foliage. The GELLO teleoperation consoles are 3D-printed kinematic replicas of the XArm7s, used for manual intervention during failure recovery. Supporting devices include an NVIDIA Orin AGX for onboard compute, a user-facing display showing the overhead camera feed over the manipulation workspace, and power and networking hardware.
The autonomous harvesting pipeline, illustrated in Figure~\ref{fig:full_overview}, proceeds as follows. Both arms first move to a home pose that provides full dual-view visibility of the harvesting workspace. Detected peppers are filtered for reachability and a target is selected; both arms then advance to a pre-grasp observation waypoint, from which a fine-grained pose estimate is obtained via YOLOv11 instance segmentation and superellipsoid fitting (Section \ref{sec:pose_estimation}). This refined pose drives parametric grasp generation, which samples candidate gripper poses along a circle orthogonal to the stem axis and selects a collision-free configuration (Section \ref{sec:grasp_planning}). Motion planning coordinates both arms using a combination of Cartesian planning and RRT* with a custom weighted optimization objective and a shared workspace division strategy that enables parallelized planning and execution (Section \ref{sec:motion_planning}). The gripper arm grasps and stabilizes the pepper while the cutter severs the peduncle, after which the gripper deposits the fruit into storage. When autonomy fails under heavy occlusion, a GELLO-based teleoperation system enables failure recovery at 100 Hz control with sub-30 ms latency (Section \ref{sec:teleoperation}).

\subsection{Bell Pepper Pose Estimation}
\label{sec:pose_estimation}
\begin{figure}[ht]
\begin{center}
        \includegraphics[trim={0, 0.7cm, 0, 0.7cm}, clip, width=\linewidth]{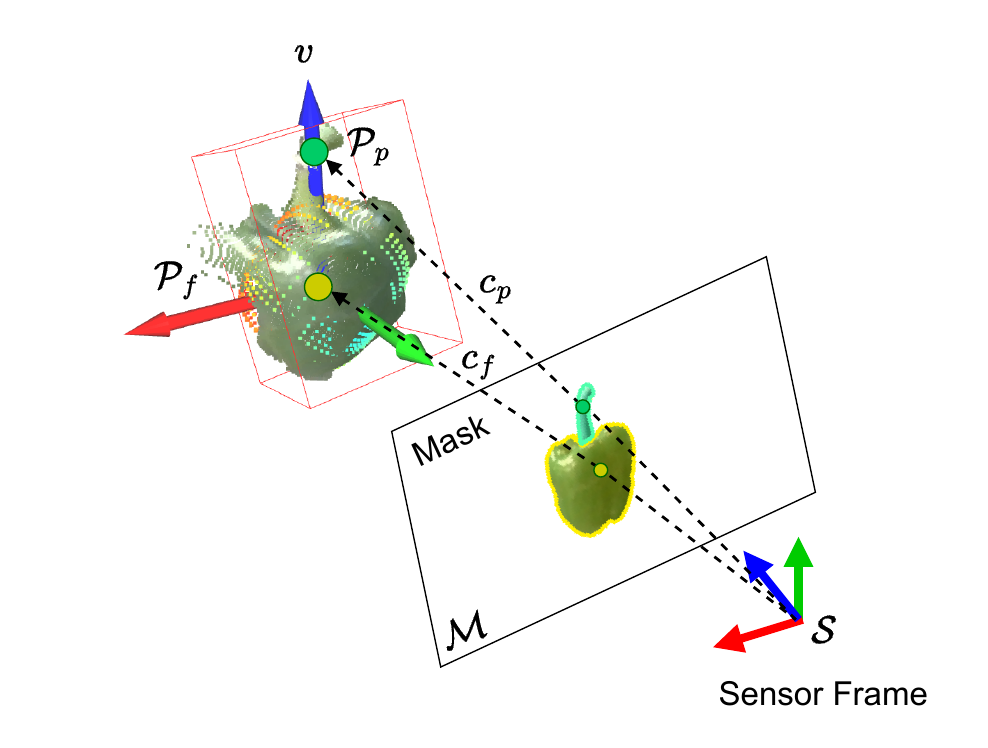}

\end{center}
        \caption{{Pepper pose estimation: We use a fitted superellipsoid to refine the position of the pepper and its orientation based on the peduncle axis.}}
        \label{fig:pose_estimation}
\end{figure}

Accurate pose estimation of bell peppers in unstructured agricultural environments is challenging due to heavy leaf occlusion, varying illumination, and the non-standard geometry of the fruit. We present a coarse-to-fine pose estimation method that integrates deep-learning-based instance segmentation with regularized superellipsoid fitting to recover the 6-DOF pose and shape parameters required for coordinated dual-arm harvesting.

The dual-RGBD camera configuration of our method enables active, multi-perspective perception that mitigates single-viewpoint occlusions, a primary failure mode in prior automated harvesting systems. During the harvesting cycle, the cameras provide complementary views that are back-projected into 3D point clouds $\mathcal{P}_{f}$ and $\mathcal{P}_{p}$ for the fruit and peduncle, respectively.

\subsubsection{Instance Segmentation}
We use two YOLOv11 \cite{yolo11_ultralytics} segmentation models fine-tuned on our dataset to obtain pepper fruit and peduncle segmentation masks. The models produce fruit and peduncle masks, that are used to extract the corresponding point clouds via depth back-projection.

\subsubsection{Coarse Pose Estimation}
The initial pose estimate is derived from the geometric relationship between the fruit body and peduncle. The coarse position is set to the fruit centroid, $t_{\mathrm{init}} = c_f$. The primary axis vector $v = c_p - c_f$ is defined from the fruit centroid to the peduncle centroid, aligning with the stem direction (see Figure \ref{fig:pose_estimation}). A local orthonormal frame $\{\hat{x}, \hat{y}, \hat{z}\}$ is then constructed as follows:
\[
  \hat{z} = \frac{v}{\|v\|}, \quad
  \hat{x} = \frac{v \times c_f}{\|v \times c_f\|}, \quad
  \hat{y} = \hat{z} \times \hat{x},
\]
where the cross product with the optical center vector $c_f$ serves as a stable reference to resolve the rotational ambiguity about the stem axis. The coarse rotation is $R_{\mathrm{init}} = [\hat{x} \mid \hat{y} \mid \hat{z}]$.

\subsubsection{Pose Refinement via Superellipsoid Fitting}
To refine the fruit position and recover shape parameters, we fit a superellipsoid to the partial fruit point cloud. 

A superellipsoid is defined by the implicit function $F(p;\Lambda) = 1$, where $\Lambda = \{a, b, c, \epsilon_1, \epsilon_2\}$ are the semi-axis lengths and curvature exponents. For a point $p = [x, y, z]^\top$ in the object's canonical frame:

\[
  F(p; \Lambda) = \left( \left|\frac{x}{a}\right|^{\frac{2}{\epsilon_2}} + \left|\frac{y}{b}\right|^{\frac{2}{\epsilon_2}} \right)^{\frac{\epsilon_2}{\epsilon_1}} + \left|\frac{z}{c}\right|^{\frac{2}{\epsilon_1}},
\]
with $0.1 < \epsilon_1, \epsilon_2 < 1.9$, allowing the representation of shapes from cubic to spherical, well-suited for modeling bell peppers. A sensor-frame point $p_s$ is transformed to the canonical frame as $p_c = R(\theta)^\top (p_s - t)$.

We optimize the full parameter set $\Theta = \{\Lambda, t, \theta\}$, comprising the shape, translation, and rotation (parameterized as a rotation vector). The per-point residual follows the cost function proposed in ~\cite{solina}:
\(
  r_i(\Theta) = \sqrt{abc}\, \left| F(p_{c,i};\Lambda)^{\epsilon_1/2} - 1 \right|,
\)
which scales the radial algebraic distance by volume to ensure uniform convergence across scales. Because RGB-D sensors observe only a partial surface, the fitting problem is inherently ill-posed. We therefore introduce regularization priors into the total cost:
\[
  E_{\mathrm{total}}(\Theta) = \sum_i r_i(\Theta)^2 + \lambda_c \|t - t_{\mathrm{init}}\|^2 + \lambda_s E_{\mathrm{scale}},
\]
where the \textit{center prior} $\|t - t_{\mathrm{init}}\|^2$ penalizes deviation from the coarse position, and the \textit{scaling isotropy prior} $E_{\mathrm{scale}} = \sqrt{(a-b)^2 + (b-c)^2 + (c-a)^2}$ favors the roughly isotropic shape characteristic of bell peppers. The non-linear least squares problem is solved using the Trust Region Reflective (TRF) algorithm.

\subsubsection{Final Pose Construction}
The superellipsoid fitting refines the fruit center position but does not directly yield the final orientation. Instead, the orientation is recomputed by substituting the refined fruit center into the coarse frame construction: the axis vector $v = c_p - t_{\mathrm{refined}}$ is recalculated from the peduncle centroid to the refined center, and the orthonormal basis $\{\hat{x}, \hat{y}, \hat{z}\}$ is rebuilt as before. This yields the final rotation $R = [\hat{x} \mid \hat{y} \mid \hat{z}]$.
% Experimental evaluation using Aruco marker ground-truth demonstrates that this refined pipeline consistently achieves a success rate of over 60 \%, even in the presence of sensor occlusions.
\subsection{Grasp Planning}
\label{sec:grasp_planning}
\begin{figure}[ht]
\begin{center}
        \includegraphics[trim={1cm, 0.6cm, 1cm, 1.1cm}, clip, width=0.95\linewidth]{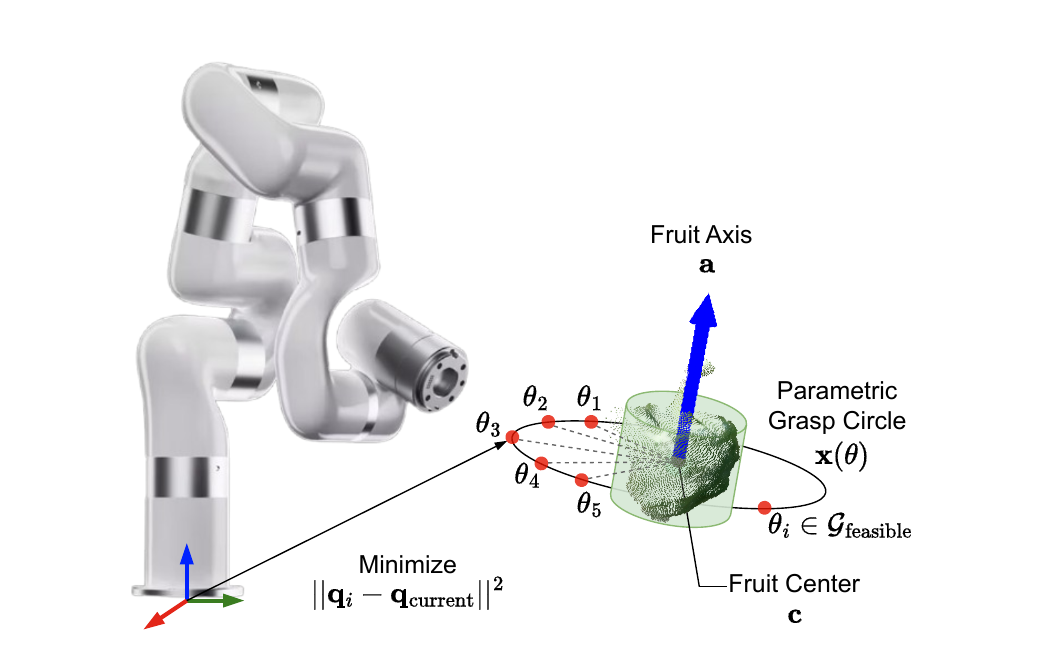}
\end{center}
        \caption{{Parametric Circle Grasp Selection: Points are sampled on the circle perpendicular to the pepper axis. The distance from the robot base for each point is minimized.}}
        \label{fig:pc}
\end{figure}

We represent the refined fruit pose as $(\mathbf{c}, \mathbf{R}) \in SE(3)$, where $\mathbf{c} \in \mathbb{R}^3$ is the fruit center and $\mathbf{R} \in SO(3)$ encodes orientation. The grasping axis aligned with the stem direction is given by $\mathbf{a} = \mathbf{R}\hat{z}$.

\paragraph{Parametric Grasp Generation.}
We construct a grasp circle (see Figure \ref{fig:pc}) in the plane orthogonal to $\mathbf{a}$ passing through $\mathbf{c}$. Let $\{\mathbf{u}_1, \mathbf{u}_2\}$ form an orthonormal basis spanning this plane with $\mathbf{u}_2 = \mathbf{a} \times \mathbf{u}_1$, as represented in Figure \ref{fig:pc}. The parametric grasp circle of radius $r$ is:
\[
\mathbf{x}(\theta) = \mathbf{c} + r(\mathbf{u}_1 \cos\theta + \mathbf{u}_2 \sin\theta), \quad \theta \in [0,2\pi).
\]
The radial approach direction at each candidate point is $\mathbf{d}(\theta) = (\mathbf{c} - \mathbf{x}(\theta))/\|\mathbf{c} - \mathbf{x}(\theta)\|$, yielding candidate end-effector poses:
\[
\mathbf{T}_g(\theta) = \begin{bmatrix} \mathbf{R}_g(\theta) & \mathbf{x}(\theta) \\ \mathbf{0}^\top & 1 \end{bmatrix},
\]
where $\mathbf{R}_g(\theta)$ orients the gripper along $\mathbf{d}(\theta)$.

\subsubsection{Grasp Selection}
We discretize the circle into $N$ candidates at $\theta_i = 2\pi i/N$ and compute inverse kinematics $\mathbf{q}_i = \text{IK}(\mathbf{T}_g(\theta_i))$ subject to joint limits and collision constraints. The feasible set is:
\[
\mathcal{G}_{\text{feasible}} = \{\theta_i \mid \text{IK}(\mathbf{T}_g(\theta_i)) \text{ exists and is collision-free}\}.
\]
We select the first kinematically feasible grasp encountered during discretization, or alternatively the grasp closest to the current configuration:
\[
\theta^* = \arg\min_{\theta_i \in \mathcal{G}_{\text{feasible}}} \|\mathbf{q}_i - \mathbf{q}_{\text{current}}\|^2.
\]
This geometry-driven approach efficiently generates collision-free grasps.

\begin{figure*}[ht]
\begin{center}
        \includegraphics[width=0.95\linewidth]{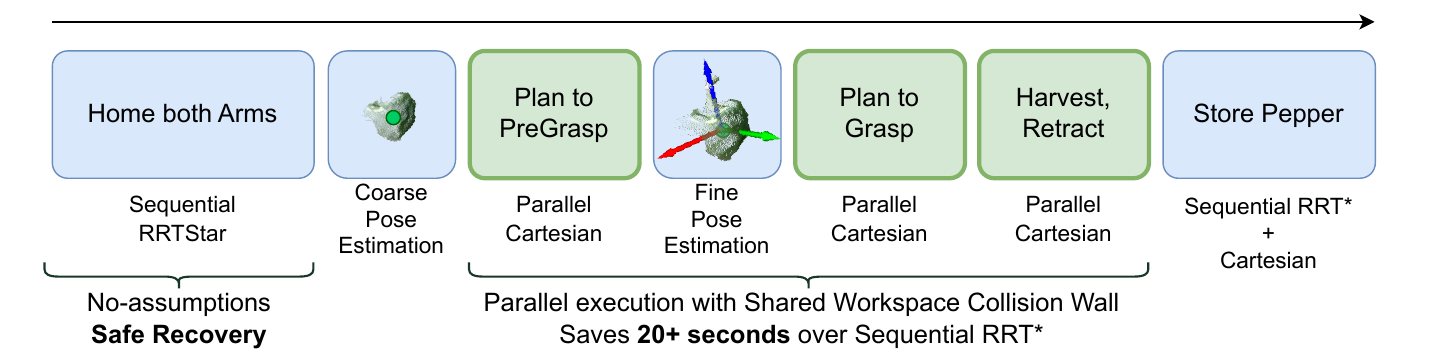}
\end{center}
        \caption{{Overall harvesting logic and planning strategies. Green highlighted motions are parallelized in terms of planning and execution.}}
        \label{fig:parallel-harvest}
\end{figure*}
\subsection{Task and Motion Planning}
\label{sec:motion_planning}

A State Machine is used to coordinate the overall harvest flow, and interfaces with a MoveIt-based motion planner to plan and execute motions on both arms, in order to guide both end effectors to the grasp positions and store the harvested fruit in the storage bin. The overall harvesting process, depicted in Figure \ref{fig:parallel-harvest}, has four main stages:

\begin{enumerate}
    \item Home Position: Both manipulators are moved to the home position, where the palm cameras are positioned to detect reachable peppers within the workspace.
    \item Move to Pre-grasp: When a set of peppers are detected and received from the Perception model, the unreachable peppers are filtered out of the list, and the pepper closest to the center of the workspace is selected for harvest. Both arms move to a pre-grasp pose approximately 20cm away from the estimated center of the fruit, with a vertical offset to help with visibility of the peduncle. A fine pose estimate of the target pepper is taken from this vantage point for greater accuracy.
    \item Move to Grasp: The fine pose estimate of the pepper is used to for Grasp Planning. The end effectors approach and harvest the target pepper.
    \item Move to Storage: The arms retract from the pepper. The cutter arm goes back to the home position while the gripper arm deposits the pepper within the storage bin, and moves back to home position.
\end{enumerate}

A combination of RRT* and Cartesian movement planning is used during the harvesting cycle. For long-range motions which need to consider complex collisions such as homing the manipulator arms or moving from harvest to storage, the system employs the RRT* algorithm. For short-range motions that require smooth task-space movement an d expect no obstacles, deterministic Cartesian planning is used to move to pre-grasp, grasp, and retraction poses. 

\begin{figure}[t]
\begin{center}
        \includegraphics[width=\linewidth]{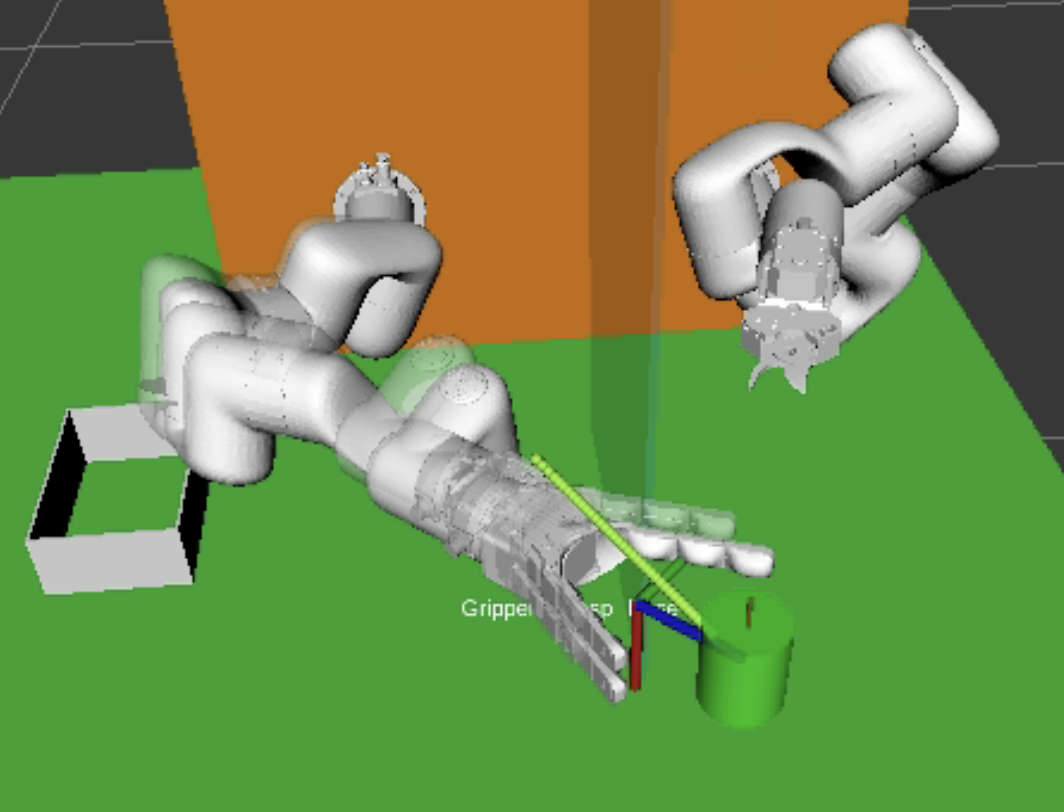}
        \caption{{The translucent blue collision between the arms divides the shared workspace, which allows for parallelized planning and execution.}}
        \label{fig:sim-workspace-div}
\end{center}
\end{figure}

\textbf{Shared Workspace Division.} 
The manipulator motion plans are computed, then executed, separately for each arm, while only assuming static obstacles and collision. Thus, during open-loop execution of the motion, any unforeseen dynamic obstacles (such as the other manipulator's movement) would render the computed plan invalid. To avoid invalidating motion plans and introduce efficient parallelization, a collision wall is introduced in the planning scene to split the shared workspace of the arms as seen in Figure \ref{fig:sim-workspace-div}. This wall is positioned such that it corresponds to the target pepper's horizontal position, such that the offline computed plans are not affected by the movement of the opposite arm; the wall is added to the planning scene after the target pepper's coarse pose is determined, and removed from the scene during the final approach to avoid planning failures when approaching the pepper pose. This allows each manipulator to safely plan within its confined task space without reasoning about the opposite manipulator pose, which results in a significantly smaller search space as an additional advantage.

\textbf{Parallelization of Planning and Execution.} 
 With the shared workspace division logic, the final implementation parallelizes the planning and execution of dual-arm actions (such as moving a harvested pepper to storage) by overlapping the execution of the first arm with the planning process for the second arm, saving approximately 20 seconds per harvest compared to sequential RRT* calls.

The pre-grasp, grasp, and retraction motions depicted in Figure \ref{fig:parallel-harvest} are parallelized in this fashion, while the initial homing and final storage motions cannot be parallelized since the shared collision wall can only be added when the arms are in a known state (e.g. at home poses).

\textbf{Weighted Optimization Objective.} The default RRT* algorithm within MoveIt employs a default optimization objective of path length within the arm's joint space. The default planner sometimes creates long-winding paths in task space—for example, a 360-degree rotation about the robot's base—that still are minimized joint space costs due to only moving a single joint. To remedy this, a custom OMPL $\mathrm{OptimizationObjective}$ is created with a weighted sum of the joint space cost and the task space end-effector distance:
\[\mathrm{motionCost}(s_a, s_b) = ||s_a - s_b||_2 + \lambda_{FK} ||\mathrm{FK}(s_a) - \mathrm{FK}(s_b)||_2\]
Where $s_a, s_b$ are arbitrary joint configuration states for a manipulator, $\mathrm{FK}$ is the forward kinematics function that outputs the end-effector pose, and $\lambda_{FK}$ is a hyperparameter that tunes the trade-off between the joint-space and task-space costs. The overall $\mathrm{motionCost}$ is evaluated during RRT* planning as the cost of the edge between two configurations.

\begin{figure}[ht]
\centering
\includegraphics[width=0.99\linewidth, trim=50 0 40 10, clip]{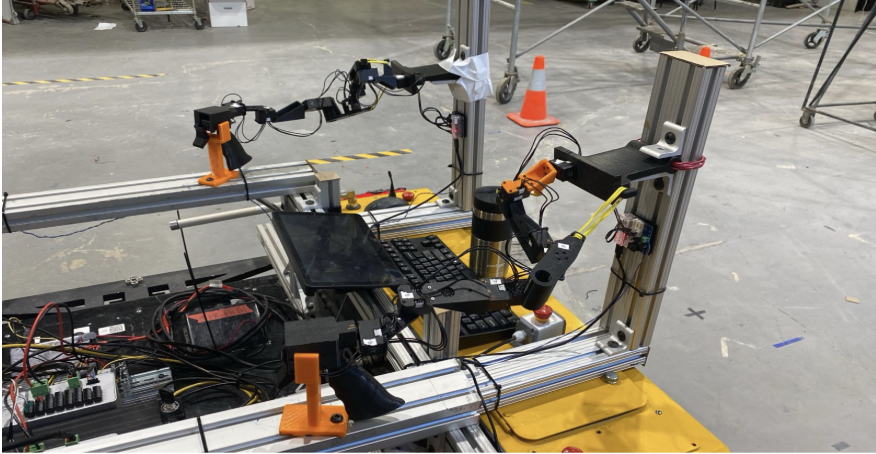}
\caption{The GELLO-based teleoperation consoles used for manual failure recovery. Each console is a 3D-printed kinematic replica of the XArm7, with Dynamixel motors emulating the 7-DOF joint states to provide intuitive, isomorphic control of the manipulators.}
\label{fig:teleop}
\end{figure}

\subsection{Teleoperation for Failure Recovery}
\label{sec:teleoperation}
Despite the robust autonomy stack, the system remains susceptible to failure modes in highly unstructured environments, particularly when dealing with heavy crop occlusion. To ensure operational continuity, we integrated a high-fidelity teleoperation system allowing for manual intervention using two XArm7 manipulators, as shown in Figure~\ref{fig:teleop}.

Our architecture builds upon the GELLO framework \cite{wu2024GELLO}, utilizing two 3D-printed consoles that share an identical kinematic configuration with the XArm-7s to provide intuitive, isomorphic control. These consoles are fabricated through a 3D printing process and are fitted with Dynamixel motors which, rather than providing active drive, are used to emulate the 7-DOF joint states.

The software backbone consists of a ZeroMQ (ZMQ) distributed messaging pattern designed for high-frequency data transmission. A ZMQ client captures high-resolution joint inputs from the teleoperation consoles, while a ZMQ server synchronizes the dual-arm movements based on these inputs. To ensure fluid motion and prevent oscillations during delicate harvesting tasks, the server's internal control loop operates at 100 Hz and the data is transmitted from the client to the server at 40 Hz, maintaining a round-trip latency of less than 30 ms. Prior to physical deployment, the entire control stack was validated within the MuJoCo physics engine to ensure trajectory fidelity, joint-to-joint mapping accuracy, and collision avoidance.

\subsection{Dataset}
\begin{figure}[ht]
    \centering
    \includegraphics[width=0.48\textwidth]{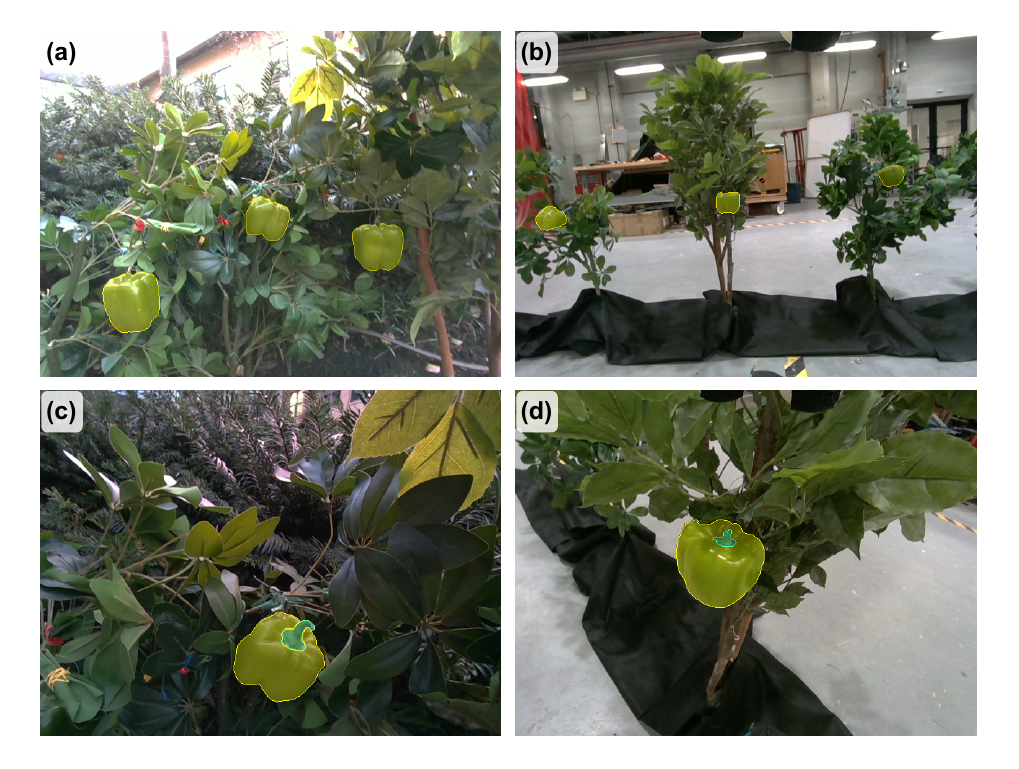}
    \caption{{Overview of our dataset - \textbf{(a)} Outdoor scene with Coarse annotations \textbf{(b)} Indoor scene with coarse annotations. \textbf{(c)} Outdoor Scene with fine annotations. \textbf{(d)} Indoor Scene with fine annotations.}}
    \label{fig:dataset}
\end{figure}

Our dataset comprises over 3,200 high-resolution images captured using an Intel RealSense D405 RGB-D sensor across both indoor and outdoor domains, with representative samples shown in Figure~\ref{fig:dataset}. For data collection, we developed a proxy for the green pepper farm by suspending green bell peppers within a row of synthetic foliage, and recorded video sequences under both indoor and outdoor conditions. The dataset employs a hierarchical annotation structure organized by domain and granularity: coarse annotations label pepper fruits only, while fine annotations include both pepper and peduncle masks. This dual-layer strategy facilitates multi-stage training for perception systems, enabling a progression from coarse object detection to the high-precision peduncle localization required for coordinated robotic cutting. The indoor subset provides a baseline of 1,033 coarse and 1,085 fine images designed to distinguish targets from robotic hardware and lab clutter, while the outdoor subset captures the environmental variability of field conditions as detailed in Table~\ref{tab:dataset_distribution}. All annotations were generated using a semi-automated pipeline based on SAM \cite{kirillov2023segment} and subsequently verified by human annotators.

\begin{table}[ht]
\centering
\small % Sets the font size for the entire table/caption
\setlength{\tabcolsep}{1pt} % Reduces horizontal space between columns (Adjust as needed)
\caption{Distribution of the VADER Dataset by Environment and Annotation Type}
\label{tab:dataset_distribution} % Must be AFTER the caption

\begin{tabular}{@{}lllc@{}}
\toprule
\textbf{Environment} & \textbf{Condition} & \textbf{Annotation Type} & \textbf{Image Count} \\ \midrule
Indoor               & Controlled         & Coarse (Fruit Only)      & 1,033                \\
Indoor               & Controlled         & Fine (Fruit + Peduncle)  & 1,085                \\
Outdoor              & Evening/Sunset     & Coarse (Fruit Only)      & 350                  \\
Outdoor              & Evening/Sunset     & Fine (Fruit + Peduncle)  & 250                  \\
Outdoor              & Sunny              & Mixed (Coarse/Fine)      & 205                  \\
Outdoor              & Shaded             & Mixed (Coarse/Fine)      & 335                  \\ \midrule
\textbf{Total}       &                    &                          & \textbf{3,258}       \\ \bottomrule
\end{tabular}
\end{table}

% \subsubsection{Data Collection Setup}
% Data acquisition was performed using an Intel RealSense D405 RGB-D sensor to capture high-resolution spatial and color information. We developed a proxy for the green pepper farm by suspending green bell peppers within a row of synthetic foliage. Using this experimental setup, we recorded a dataset comprising indoor and outdoor video sequences. We use a semi automated pipeline for annotating the bell peppers and peduncles in the image, using SAM \cite{kirillov2023segment}.
\section{EXPERIMENTAL RESULTS}

This section presents a series of experiments evaluating the proposed autonomous pepper harvesting pipeline across segmentation, motion planning in simulation, and real-world validation. Evaluations were conducted both in simulation and on the physical VADER system to systematically assess perception accuracy, reachability, and harvest success rate. All experiments were performed under controlled lighting and background conditions to ensure consistency across trials.

\subsection{Pepper and Peduncle Segmentation}
Table~\ref{table_segmentation_results} summarizes the segmentation performance of our fine-tuned YOLOv11m-seg models. The pepper model, pretrained on the BUP20 \cite{smitt2023pag} and Kaggle \cite{montoya2021sweet} datasets and fine-tuned on our custom dataset, achieves strong detection performance with an F1 score of 0.942 and mAP@50 of 0.97. The peduncle model similarly achieves reliable detections, with an F1 score of 0.832 and mAP@50 of 0.884. Both models run at under 25\,ms per frame on an NVIDIA RTX 4080, supporting real-time perception at 60\,Hz. Compared to \cite{lenz2024hortibot}, who report a peduncle segmentation mAP@50 of 0.741 using YOLOv8, our approach improves performance through the use of YOLOv11, a larger combined training dataset, and an improved training strategy. The relatively lower peduncle mAP@50-95 of 0.388 indicates that while peduncles are reliably detected, precise mask localization remains an open challenge: Improving this metric is a key direction for future work, as tighter peduncle masks directly benefit cutting accuracy.

\begin{table}[ht]
\caption{{Pepper and Peduncle Segmentation Results}}
\label{table_segmentation_results}
\begin{center}
\begin{tabular}{|c|c|c|c|c|}
\hline
Metric & F-1 score & mAP50 & mAP50-95 & Inference speed (ms)\\
\hline
Pepper & 0.942 & 0.97 & 0.91 & 15.15\\
\hline
Peduncle & 0.832 & 0.884 & 0.388 & 22.11\\
\hline
\end{tabular}
\end{center}
\end{table}

\subsection{Reachability Simulation Results}
In order to decouple system development from hardware availability, a simulated instance of the system is created in Gazebo Classic, with simulated detachable goal peppers that are attached to the peduncle with a force-sensitive fixed joint. The ground truth pose of the pepper is injected with Gaussian noise to create coarse and fine pose estimates, which are used by the State Machine and Motion Planner to approach and harvest the pepper in the simulator. This enables integration testing of the motion planning across thousands of simulated harvests to discover constraints on reachability due to the limited reach of the XArm manipulators, and test the system thoroughly before full deployment.

% \begin{table}[ht]
% \centering
% \renewcommand{\arraystretch}{1.3} % increase row height
% \setlength{\tabcolsep}{10pt} % increase column spacing
% \begin{tabular}{|l|c|}
% \hline
% \textbf{Phase} & \textbf{Time (s)} \\
% \hline
% Coarse Pose & 9.54 \\
% Fine Pose   & 2.68 \\
% Grasp       & 2.35 \\
% Storage     & 35.23 \\
% \hline
% \end{tabular}
% \caption{Execution time for each phase of the system.}
% \label{tab:phase_times}
% \end{table}

\begin{figure}[ht]
    \centering
    \includegraphics[width=0.45\textwidth]{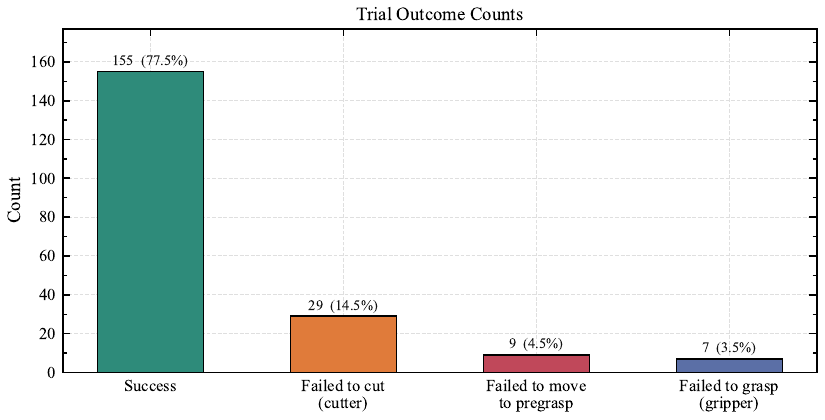}
    \caption{{Success and failure metrics for simulated harvesting}}
    \label{fig:sim_results}
\end{figure}

A simulation director program is used to uniformly randomize pepper pose as tuples of $\{x, y, z, pitch, roll\}$ within a workspace bounding box and with up to 0.5 radians of combined tilt (pitch and tilt) from a neutral orientation. As shown in Figure \ref{fig:sim_results}, over 100 such randomized runs, peaks of gripper and cutter failures are observed at either ends of the workspace as the pepper pose is less reachable to the opposing arm. Statistically, to maximize reachability, the targeted pepper must be within 10cm of the center of the robot platform horizontally.

% \begin{figure}[H]
% \begin{center}
%         \includegraphics[width=\linewidth]{images/sim-results.png}
% \end{center}
%         \label{fig:sim_results}
%         \caption{Success and Failure metrics for simulated harvesting, binned by horizontal pepper position in steps of 5cm.}
% \end{figure}

%We can use one of these two figs, check which one is best

% \begin{figure}[H]
%     \centering
%     \includegraphics[width=0.5\textwidth]{images/vader_real1.png}
%     \caption{{Success and failure metrics for simulated harvesting, binned by horizontal pepper position}}
%     \label{fig:sim_results}
% \end{figure}

% \begin{figure}[H]
%     \centering
%     \includegraphics[width=0.5\textwidth]{images/vader_real2.png}
%     \caption{{Success and failure metrics for simulated harvesting, binned by horizontal pepper position.}}
%     \label{fig:sim_results}
% \end{figure}

% \subsection{Harvesting Results}

\subsection{Harvesting Results}

\textbf{In simulation:} The system achieved a mean harvest success rate of 92.3\% in all trials. The average harvest cycle time was 69.25 s, of which 14.4 s corresponded to visual perception and motion planning and 35.23 s to cutting action and storage of the pepper. Real-world trends mirrored simulation insights, confirming reachability and pose accuracy and a full state-machine cycle as the main determinants of success.

\begin{figure}[!htbp]
    \centering
    \includegraphics[width=0.49\textwidth]{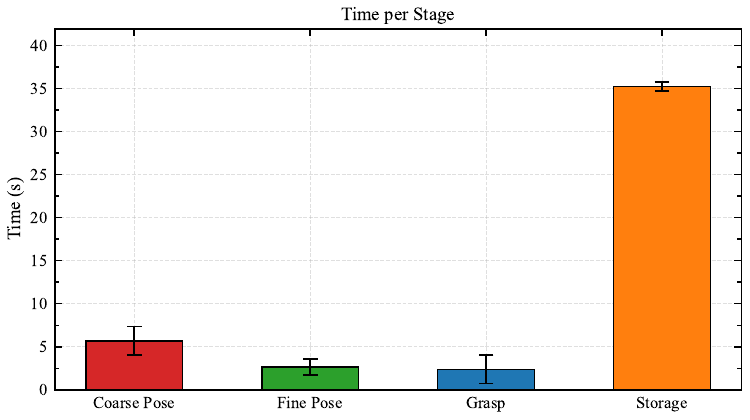}
    \caption{{Execution time for each phase of the system on the testbed.}}
    \label{fig:harvesting_setup}
\end{figure}
\textbf{On testbed:} We evaluated the full VADER platform on an indoor testbed consisting of a row of six artificial plants with real green bell peppers, arranged under pose distributions similar to those used in simulation. A total of 47 harvesting trials were executed with randomized pepper placements and varying lighting conditions. Our system achieved 30 successful harvests out of 47 trials, yielding a success rate of 61\%. The overall harvest time averaged across the trials was 69.25s. Figure~\ref{fig:harvesting_setup} shows the mean execution time for each stage in the harvesting process.
% \subsubsection{Failure Mode Analysis} Tom: changing to bolded text for consistenc
\\
\textbf{Failure Mode Analysis:} Of the 17 failed trials, the most common failure mode was a cutter offset error (13 out of 17 cases), where the cutter arm missed the peduncle by a small vertical offset. This was particularly pronounced in our indoor mockup, because the bell peppers were suspended by strings, requiring precise calibration. The remaining four failures were distributed across three other modes: Storage failure where the motion planner timed out in cartesian mode before moving to storage, a grasp failure where the gripper did not produce a deterministic grasp, and a pose estimation failure where noisy depth data caused the system to estimate the pepper's position outside the reachable workspace.

\section{CONCLUSION}

This work presented VADER, a dual-arm mobile manipulation system for autonomous bell pepper harvesting that addresses the labor intensity, cost, and accessibility barriers of current agricultural automation. By decoupling grasping and cutting across two coordinated 7-DOF arms mounted on a mobile platform, VADER advances beyond greenhouse-centric solutions toward practical open-field deployment.
A key contribution is our annotated bell pepper dataset spanning indoor and outdoor environments with fruit and peduncle labels, providing a benchmark for detection and segmentation under realistic conditions including occlusion, lighting variation, and scene clutter. Building on this, we present an end-to-end harvesting pipeline integrating superellipsoid-based pose estimation, parametric grasp planning, parallelized dual-arm motion planning, and GELLO-based teleoperation for failure recovery. Indoor experiments demonstrate reliable grasp-and-cut performance with minimal plant damage, validating our core design principle that autonomy and human oversight are complementary rather than competing. Future work will focus on three directions: (i) full integration of autonomous platform navigation for row traversal between plants, (ii) outdoor field testing to evaluate robustness under diverse environmental conditions, and (iii) active perception strategies to improve reliability under partial fruit occlusion. Together, these advances will move VADER toward practical deployment as a complete field harvesting solution.
% This work introduced VADER, a dual-arm mobile manipulation system for autonomous bell pepper harvesting in unstructured fields, addressing the labor, cost, and accessibility challenges in bell pepper harvesting. By decoupling grasping and cutting across two coordinated 7-DOF arms on a mobile platform, the system narrows the gap between greenhouse-centric solutions and open-field needs.

% We also contribute a structured bell pepper dataset spanning indoor and outdoor domains with annotations for both fruit and peduncle, providing a foundation to develop and benchmark detection and segmentation methods under realistic lighting, occlusion, and scene variability.

% Using this dataset, we demonstrate an end-to-end dual-arm harvesting framework that combines learned perception, coordinated motion planning, and a GELLO-based teleoperation system. The experimental results, including high indoor success rates, show that the architecture enables reliable grasp-and-cut behavior while preserving the plant and produce. The teleoperation layer further supports our design choice to treat autonomy and human oversight as complementary, enabling safe recovery from failure modes that remain difficult for full autonomy.

% Looking ahead, we plan to (i) fully integrate control of the mobile platform to enable autonomous navigation between plants, (ii) perform testing of our system on outdoor environments to check robustness against diverse conditions, and (iii) perform additional active perception to increase robustness against partial fruit occlusions.

\addtolength{\textheight}{-12cm}   % This command serves to balance the column lengths

%%%%%%%%%%%%%%%%%%%%%%%%%%%%%%%%%%%%%%%%%%%%%%%%%%%%%%%%%%%%%%%%%%%%%%%%%%%%%%%%

%%%%%%%%%%%%%%%%%%%%%%%%%%%%%%%%%%%%%%%%%%%%%%%%%%%%%%%%%%%%%%%%%%%%%%%%%%%%%%%%

\section*{ACKNOWLEDGMENT}

The authors would like to express their sincere gratitude to Dr. George Kantor for providing access and support for the Warthog robot and one of the XArm manipulators, which were instrumental in enabling the real-world experiments. 

The team also gratefully acknowledges the guidance and continuous supervision of Dr. John M. Dolan and Dr. Dimitrios during the MRSD Masters Program, whose insights and direction were critical in shaping the methodology and executing this work successfully.
%%%%%%%%%%%%%%%%%%%%%%%%%%%%%%%%%%%%%%%%%%%%%%%%%%%%%%%%%%%%%%%%%%%%%%%%%

\bibliographystyle{IEEEtran} % Or 'plain', 'alpha', etc., depending on your required style
\bibliography{references}

\end{document}